\documentclass{article}

\usepackage{arxiv}

\usepackage[T1]{fontenc}    
\usepackage{hyperref}       
\usepackage{url}            
\usepackage{booktabs}       
\usepackage{amsfonts}       
\usepackage{nicefrac}       
\usepackage{microtype}      
\usepackage{lipsum}         
\usepackage{graphicx}
\usepackage[numbers]{natbib}
\usepackage{doi}

\usepackage{listings}
\usepackage{listingsutf8} 

\usepackage{amsmath} 
\usepackage{xcolor}
\usepackage{color}

\usepackage{cleveref}       
\usepackage{tikz}
\usepackage{xspace}

\usepackage{amsmath,amssymb}
\usepackage{alltt}

\definecolor{keywordcolor}{rgb}{0.0,0.0,0.6}
\definecolor{codegreen}{rgb}{0,0.6,0}
\definecolor{codegray}{rgb}{0.5,0.5,0.5}
\definecolor{codepurple}{rgb}{0.58,0,0.82}
\definecolor{backcolour}{rgb}{0.95,0.95,0.92}

\lstdefinestyle{mystyle}{%
    backgroundcolor=\color{backcolour},   
    commentstyle=\color{codegreen},
    keywordstyle=\color{keywordcolor},
    numberstyle=\tiny\color{codegray},
    stringstyle=\color{codepurple},
    basicstyle=\ttfamily\footnotesize,
    breakatwhitespace=false,         
    breaklines=true,                 
    captionpos=b,                    
    keepspaces=true,                 
    numbers=left,                    
    numbersep=5pt,                  
    showspaces=false,                
    showstringspaces=false,
    showtabs=false,                  
    tabsize=2,
    morekeywords={assert},
}

\definecolor{light-gray}{gray}{0.95}

\definecolor{kwgreen}{HTML}{3EAE2B}
\definecolor{kwblue}{HTML}{0068C7}
\definecolor{kworange}{HTML}{EF7724}
\definecolor{kwred}{HTML}{F42836}
\definecolor{kwdarkgreen}{HTML}{2E5524}
\definecolor{kwdarkblue}{HTML}{003765}

\newcommand{\ie}{i.e.\@\xspace}

\newcommand{\TP}[1]{\textcolor{kwgreen}{\texttt{\textbf{TP}}}}
\newcommand{\FP}[1]{\textcolor{kwred}{\texttt{\textbf{FP}}}}
\newcommand{\FN}[1]{\textcolor{kworange}{\texttt{\textbf{FN}}}}
\newcommand{\TN}[1]{\textcolor{kwblue}{\texttt{\textbf{TN}}}}

\newcommand{\mata}[1]{\textcolor{kwgreen}{\texttt{\textbf{a}}}}
\newcommand{\matb}[1]{\textcolor{kwred}{\texttt{\textbf{b}}}}
\newcommand{\matc}[1]{\textcolor{kworange}{\texttt{\textbf{c}}}}
\newcommand{\matd}[1]{\textcolor{kwblue}{\texttt{\textbf{d}}}}

\newcommand{\PPV}[0]{\texttt{\textbf{PPV}}}
\newcommand{\TPR}[0]{\texttt{\textbf{TPR}}}
\newcommand{\MCC}[0]{\texttt{\textbf{MCC}}}
\newcommand{\Fowlkes}[0]{\texttt{\textbf{FM}}}
\newcommand{\Fone}[0]{\texttt{\textbf{F1}}}

\title{%
    The MCC approaches the geometric mean of precision and recall as true negatives approach infinity.
}

\author{\href{https://orcid.org/0009-0008-8455-7514}{\includegraphics[scale=0.06]{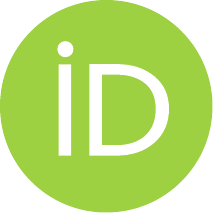}\hspace{1mm}Jon Crall}\thanks{\href{https://github.com/Erotemic}{https://github.com/Erotemic} \hspace{10pt} \texttt{erotemic@gmail.com}} \\
	Kitware Inc.\\
	\texttt{jon.crall@kitware.com}
}

\renewcommand{\shorttitle}{\textit{arXiv} Template}

\hypersetup{%
pdftitle={The MCC approaches the geometric mean of precision and recall as true negatives approach infinity.},
pdfsubject={q-cs.CV},
pdfauthor={Jon Crall},
pdfkeywords={Confusion Matrix, Binary Classification, Fowlkes--Mallows Index, Matthews Correlation Coefficient, F1, Phi-Coefficient, Ochiai Index},
colorlinks=true,       
}

\begin{document}
\maketitle

\begin{abstract}

    The performance of a binary classifier is described by a confusion matrix with four entries:
    the number of true positives (\TP{}), true negatives (\TN{}), false positives (\FP{}), and false
      negatives (\FN{}).
    The Matthews Correlation Coefficient (MCC), F1, and Fowlkes--Mallows (FM) scores are scalars that
      summarize a confusion matrix.
    Both the F1 and FM scores are based on only three of the four entries in a confusion matrix (they ignore
      \TN{}).
    Unlike F1 and FM, the MCC depends on all four entries of the confusion matrix, which can make it
      attractive in some cases.

    However, in some open world settings, measuring the number of true negatives is not straightforward.
    Object detection is such a case because the number of candidate negative boxes is effectively unbounded.
    This motivates the question:
    what is the limit of the MCC as the number of true negatives tends to infinity?

    Put plainly, as the true negative count grows, the MCC converges to the FM score, which is the geometric
      mean of precision and recall.
    This result was previously noted in the ecology literature in terms of the phi-coefficient and the
      Ochiai index, but we discuss it in the context of binary classifiers.
    Furthermore, we provide a full proof of the result, including a Lean formalization.
    We also briefly comment on the emerging role of LLMs in proof assistance and in locating prior work.
\end{abstract}

\keywords{Confusion Matrix \and Binary Classification \and Fowlkes--Mallows Index \and Matthews Correlation Coefficient \and F1 \and Phi-Coefficient \and Ochiai Index}

\section{Introduction}

Evaluation of binary classifiers is central to the quantitative analysis of
machine learning models~\cite{powers_evaluation_2011}.
Given a finite set of examples with known real labels, the quality of a set of
corresponding predicted labels can be quantified using a $2 \times 2$ confusion
matrix.  A confusion matrix counts the number of true positives (\TP{}), true
negatives (\TN{}), false positives (\FP{}), and false negatives (\FN{}) a model
predicts with respect to the real labels. A confusion matrix is written as:

\begin{equation}
\begin{bmatrix}
    \TP{} & \FP{} \\
    \FN{} & \TN{} 
\end{bmatrix}	
\end{equation}

This matrix provides a holistic view of classifier quality, however, it is
often desirable to summarize performance using fewer numbers. Two popular
metrics defined on a confusion matrix are precision and recall.

Precision --- also known as the positive-predictive-value (PPV) --- is the
fraction of positive predictions that are correct.

\begin{equation}
    \PPV{} = \frac{\TP{}}{\TP{} + \FP{}}
\end{equation}

Recall --- also known as the true positive rate (TPR), sensitivity, or
probability of detection (PD) --- is the fraction of real positive cases that
are correct.

\begin{equation}
    \TPR{} = \frac{\TP{}}{\TP{} + \FN{}}
\end{equation}

One of the most popular confusion metrics is the F1 score. 
It can be defined as the harmonic mean of precision and recall.

\begin{equation}\label{eq:fone}
    \Fone{} = \frac{2 \PPV{} \cdot \TPR{}}{\PPV{} + \TPR{}} = \frac{2 \TP{}}{2 \TP{} + \FP{} + \FN{}}
\end{equation}

A similar, but less used metric is the Fowlkes-Mallows index~\cite{fowlkes_method_1983}, which was
  originally developed for measuring the similarity between two clusterings of a set of points.
It can be defined as the geometric mean of precision and recall~\cite{tharwat_classification_2020}.

\begin{equation}\label{eq:fm}
    \Fowlkes{} = \sqrt{\PPV{} \cdot \TPR{}} = \sqrt{\frac{\TP{}}{\TP{} + \FP{}} \frac{\TP{}}{\TP{} + \FN{}}}
\end{equation}

The Matthews Correlation Coefficient (MCC)~\cite{matthews_comparison_1975} is
the Pearson correlation between the truth and predictions. It accounts for all
four terms in the confusion matrix and is defined as:

\begin{equation}\label{eq:mcc}
    \MCC{} = \frac{%
        \TP{} \cdot \TN{} - \FP{} \cdot \FN{}
    }
    {\sqrt{%
        (\TP{} + \FP{}) (\TP{} + \FN{}) (\TN{} + \FP{}) (\TN{} + \FN{})
    }}
\end{equation}

While the MCC can be a desirable measure, it requires that the number of true negatives is measurable.
In the case of object detection problems~\cite{zou2023object}, this is often intractable as the number of
  predicted boxes and missed true boxes is dwarfed by the total number of boxes that the system correctly did
  not predict.
One can see this by considering the set of all $N\times M$ boxes centered at each pixel, most of which will
  be considered true negatives.
If the width and height of the boxes are allowed to extend outside the image, then the number of predictable
  boxes is unbounded (and even if they are constrained to the bounds of the image, there will still be a very
  large number of them in real world cases).

Because calculating the number of true negatives is difficult for open-world problems like object detection,
  it is conceptually simpler to ignore true negatives and simply focus on the much smaller set of true
  positives, false positives, and false negatives, which can be used to compute PPV, TPR, F1, and FM.
While these measures have proven themselves to be effective, simply ignoring true negatives is somewhat
  unsatisfying.
If we insist on understanding how the MCC behaves in this case, we can begin by studying its limiting
  behavior.
Notice that in these open-world problems the number of true negatives is so large it is effectively infinite
  and thus we ask the question:
what happens to the MCC as the number of true negatives approaches infinity?

We find that the MCC converges to FM as the number of true negatives approaches infinity.
This corresponds to a previous observation in~\cite{decaceres_assessing_species_2008}, which we discuss in
  \Cref{sec:related_work}.
We then prove $\lim_{\TN{} \to \infty} \MCC{} = \Fowlkes$ in \Cref{sec:prose_proof}.
We strengthen this proof by providing a Lean~4 formalization in~\Cref{sec:lean}.
Finally, in \Cref{sec:ai_usage} we briefly discuss the role of LLMs in searching related work and producing
  the machine formalization.

\section{Related Work}\label{sec:related_work}

In~\cite{powers_evaluation_2011}, Powers notes that the F1 score (and consequentially any metric that only
  includes precision and recall) only takes into account three of the four measures in a confusion matrix.
Powers introduces modifications of precision and recall he refers to as informedness and markedness.
Additionally he advocates for the use of the MCC over the F1 measure.

\newcommand{\MAT}[1]{\ensuremath{
    \left[ \begin{smallmatrix} #1
    \end{smallmatrix} \right] }}

In ecology, $2\times 2$ contingency tables are commonly used to measure association/similarity between two
  binary variables.
For example, consider measuring how commonly two different species, $A$ and $B$, are observed together
  across a collection of observations (e.g., sampling sites).
At each observation, record whether $A$ is present and whether $B$ is present, yielding two binary
  variables.
The entries of the resulting $2\times 2$ table $\MAT{\mata{} & \matb{}\\ \matc{} & \matd{}}$ correspond to
  the number of observations in each joint outcome.
Specifically, the table entries are the number of observations where:
$\mata{}$:
both $A$ and $B$ are present, $\matb{}$:
$A$ is present but $B$ is absent, $\matc{}$:
$B$ is present but $A$ is absent, and $\matd{}$:
neither $A$ nor $B$ is present (a ``double zero'').
The phi coefficient is the Pearson correlation between these two binary variables.

A binary confusion matrix can similarly be viewed as a $2\times 2$ contingency table, where from a machine
  learning lens, the variables correspond to the presence / absence of a prediction and the presence / absence
  of truth.
Under the identification $(\mata{},\matb{},\matc{},\matd{})=(\TP{},\FP{},\FN{},\TN{})$, the phi coefficient
  (or $\phi$ coefficient) is identical to the Matthews correlation coefficient (MCC).
Similarly, the Ochiai index~\cite{ochiai_zoogeographic_studies_1957} is exactly the FM index.

The fact that the limit of the phi coefficient as \matd{} tends towards infinity converges to the Ochiai
  index was noted in~\cite{janson_measures_ecological_1981}.
The authors remark that the relationship is easy to see, but omit the proof.
This observation is also noted in~\cite{decaceres_assessing_species_2008,caceres_associations_species_2009}.
In an ecological context, it is desirable to derive measures that are insensitive to inclusions of
  irrelevant ``double zeros'' because:
``a comparison between meadow flowers should not be influenced by the addition of localities from swamp and
  city areas''~\cite{janson_measures_ecological_1981}.

In open-world object detection, an analogous issue arises because the set of ``negative'' instances is not a
  fixed finite collection, but an implicit (and potentially enormous) universe of candidate bounding boxes.
If one were to score a detector using the MCC, one would need to define and count \TN{} over this universe,
  which is generally ill-defined and dependent on how the candidate space is discretized.
Taking the limit $\TN{}\to\infty$ yields a stable, \TN{}-free expression, enabling one to relate properties
  of the MCC to a quantity computable from \TP{}, \FP{}, and \FN{} alone.
However, the purpose of this paper is to explore this mathematical relationship, not to prescribe evaluation
  desiderata for object detection.

While the proof is straightforward, it is not entirely trivial.
Our contribution is to note the interpretation of this limit in object detection, prove it explicitly in
  standard notation, and formalize the result in Lean 4.

\section{The Relationship Between MCC and FM}\label{sec:prose_proof}

\paragraph{Taking the limit of the MCC}

Let $\TP{}$, $\FP{}$, $\FN{}$, and $\TN{}$ be non-negative real numbers with
$\TP{} + \FP{} > 0$ and $\TP{} + \FN{} > 0$. 
The MCC and FM are defined in~\cref{eq:mcc} and~\cref{eq:fm}.

Consider the limit of the MCC as the number of true negatives approaches
infinity.

\begin{equation}
    \lim_{\TN{} \to \infty} \MCC{} = \lim_{\TN{} \to \infty}
    \frac{%
        \TP{} \cdot \TN{} - \FP{} \cdot \FN{}
    }
    {\sqrt{%
        (\TP{} + \FP{}) (\TP{} + \FN{}) (\TN{} + \FP{}) (\TN{} + \FN{})
    }}
\end{equation}

We can take this limit by applying some algebra to the body of the limit. We multiply the numerator and denominator by $\frac{1}{\TN{}}$:

\begin{equation}\label{eq:lee_step1}
    = \lim_{\TN{} \to \infty}
    \frac{%
        \frac{1}{\TN{}} (\TP{} \cdot \TN{} - \FP{} \cdot \FN{}) 
    }
    {\frac{1}{\TN{}} \sqrt{%
        (\TP{} + \FP{}) (\TP{} + \FN{}) (\TN{} + \FP{}) (\TN{} + \FN{})
    }} 
\end{equation}

We distribute the $\frac{1}{\TN{}}$ term in the numerator and denominator:

\begin{equation}\label{eq:lee_step2}
    = \lim_{\TN{} \to \infty}
    \frac{%
        (\TP{} - \FP{} \cdot \frac{\FN{}}{\TN{}}) 
    }
    {\sqrt{%
        (\TP{} + \FP{}) (\TP{} + \FN{}) (\frac{\TN{} + \FP{}}{\TN{}}) (\frac{\TN{} + \FN{}}{\TN{}})
    }}
\end{equation}

The $\frac{\TN{}}{\TN{}}$ terms in the denominator cancel:

\begin{equation}\label{eq:lee_step3}
    = \lim_{\TN{} \to \infty}
    \frac{%
        (\TP{} - \FP{} \cdot \frac{\FN{}}{\TN{}}) 
    }
    {\sqrt{%
        (\TP{} + \FP{}) (\TP{} + \FN{}) (1 + \frac{\FP{}}{\TN{}}) (1 + \frac{\FN{}}{\TN{}})
    }}
\end{equation}

The terms involving $\TN{}$ are fractions of simple rational polynomials (w.r.t.
$\TN{}$) and in each case the degree of the denominator is greater than that of the numerator, so in the
  limit each of these terms simplifies to $0$.
The equation then simplifies to:

\begin{equation}\label{eq:postlimit}
    = 
    \frac{%
        (\TP{} - \FP{} \cdot 0) 
    }
    {\sqrt{%
        (\TP{} + \FP{}) (\TP{} + \FN{}) (1 + 0) (1 + 0)
    }}
\end{equation}

Thus we find that the limit of
the MCC as true negatives approach infinity is:

\begin{equation}
    = 
    \frac{\TP{}}
    {\sqrt{%
        (\TP{} + \FP{}) (\TP{} + \FN{}) 
    }}
\end{equation}

\paragraph{Rearranging the FM}

Now, rearranging the equation for FM, we find it is equivalent to the limit of the MCC as the number of true
  negatives approaches infinity.

\begin{align}
    \Fowlkes{} &= \sqrt{\frac{\TP{}}{\TP{} + \FP{}} \frac{\TP{}}{\TP{} + \FN{}}} \\
               &= \sqrt{\frac{\TP{}^2}{(\TP{} + \FP{}) (\TP{} + \FN{})}} \\
               &= \frac{\TP{}}{\sqrt{(\TP{} + \FP{}) (\TP{} + \FN{})}} \\
               &= \lim_{\TN{} \to \infty} \MCC{}
\end{align}

Thus we have proven $\lim_{\TN{} \to \infty} \MCC{} = \Fowlkes{}$ \hspace{10pt}\dots{} or have we?
This particular argument is simple to verify, but it does assume knowledge about algebra and limits and
  assumes justifications are always clear for each step.
Perhaps a skeptic would be more convinced with a machine checkable claim.

\paragraph{Verifying the proof}

The correctness of these claims can be verified using SymPy~\cite{sympy17}.
We define a symbolic expression for the definition of the MCC and FM score.
We then use SymPy to determine the limit of the MCC as $\TN{} \to \infty$.
Finally we subtract expressions that we claim are equal, which will result in zero only if they are equal.

\begin{lstlisting}[language=Python]
from sympy import sqrt, symbols, simplify
from sympy.series import limit

tp, tn, fp, fn = symbols("tp tn fp fn", integer=True, negative=False)

# The definition of the MCC
numer = (tp * tn - fp * fn)
denom = sqrt((tp + fp) * (tp + fn) * (tn + fp) * (tn + fn))
mcc = numer / denom

# The definition of FM
FM = sqrt((tp / (tp + fn)) * (tp / (tp + fp)))

# Compute the limit of the MCC definition
mcc_lim = limit(mcc, tn, float("inf"))

# We claim the limit of the MCC and the FM are equivalent to:
mcc_lim_claim = tp / sqrt((tp + fn) * ((tp + fp)))

# Check the claim is equal to FM
assert simplify(FM - mcc_lim_claim) == 0
# Check the claim is equal to the MCC limit
assert simplify(mcc_lim - mcc_lim_claim) == 0

\end{lstlisting}

The above program does not raise an \texttt{AssertionError}, thus we have proven $\lim_{\TN{} \to \infty} \MCC{} =
  \Fowlkes{}$\hspace{10pt}\dots{} or have we?
Do we trust that SymPy's symbolic manipulations are always valid?
It is a large software package, which can have bugs, and a full audit is difficult.
However there is a proof assistant system that is designed to have a small auditable kernel:
Lean~4~\cite{moura2021lean}.
While bugs have been found in the kernel, they are actively searched for, considered high priority, and
  fixed quickly~\cite{carneiro2025lean4leanverifyingtypecheckerlean}.

\section{Formal Verification with Lean}\label{sec:lean}

We formalize the main result in Lean~4~\cite{moura2021lean}, an interactive theorem prover with a small
  trusted kernel.
A Lean proof provides a machine-checked guarantee that each algebraic manipulation and limit argument used
  in the informal derivation is valid under explicit assumptions of kernel soundness.

The proof is small enough that we can present it here, heavily commented to make the structure clear for
  readers who are unfamiliar with Lean.
It is possible to dramatically simplify the proof, but this variant is better aligned with the arguments in
  \Cref{sec:prose_proof} and more suitable as an introduction to Lean.


\noindent

\definecolor{commentcolor}{rgb}{0.0,0.5,0.0}
\definecolor{stringcolor}{rgb}{0.58,0.0,0.82}
\definecolor{numbercolor}{rgb}{0.5,0.5,0.5}
\definecolor{tacticcolor}{rgb}{0.58,0.0,0.82}

\lstdefinelanguage{lean}{%
  sensitive=true,
  morecomment=[l]{--},
  morecomment=[s]{/-}{-/},
  morekeywords=[1]{import,open,namespace,section,end,noncomputable,variable,variables,
    theorem,lemma,def,abbrev,example,instance,structure,class,inductive,notation,
    by,where,let,have,show,fun,match,with,if,then,else,do,return,in,forall},
  morekeywords=[2]{simp,simpa,rw,erw,dsimp,unfold,ring,ring_nf,nlinarith,linarith,
    positivity,norm_num,field_simp,omega,tauto,exact,refine,apply,assumption,
    intro,intros,ext,funext,cases,rcases,obtain,constructor,left,right,
    classical,by_contra,contrapose,finish,taut,done,convert,using,suffices,calc},
}

\newcommand{\surdSym}{\ensuremath{\raisebox{0.3ex}{\scalebox{0.8}{$\surd$}}}}

\lstdefinestyle{leanpretty}{%
    language=lean,
    columns=fullflexible,
    backgroundcolor=\color{backcolour},   
    commentstyle=\color{codegreen},
    keywordstyle=\color{keywordcolor},
    numberstyle=\tiny\color{codegray},
    stringstyle=\color{codepurple},
    basicstyle=\ttfamily\footnotesize,
    breakatwhitespace=false,         
    breaklines=true,                 
    captionpos=b,                    
    keepspaces=true,                 
    numbers=left,                    
    numbersep=5pt,                  
    showspaces=false,                
    showstringspaces=false,
    showtabs=false,                  
    tabsize=2,
    literate=
      {`}{\textasciigrave}1
      {\\Real}{$\mathbb{R}$}5
      {\\le}{$\le$}3
      {\\ge}{$\ge$}3
      {\\ne}{$\ne$}3
      {\\to}{$\to$}3
      {\\in}{$\in$}3
      {\\nhdsSym}{\ensuremath{\mathcal{N}}}8
      {\\Infinity}{\ensuremath{\infty}}9
      {\\root}{\surdSym}5
      {\\eqF}{\ensuremath{=\raisebox{-0.3ex}{$^{\mathrm{f}}$}}}4
}

\lstset{style=leanpretty}

First we import basic mathlib functionality and define the terms used in the proof.

\begin{lstlisting}[language=lean]
import Mathlib.Tactic
open Filter Topology

noncomputable section

/-- Precision (positive predictive value) -/
def PPV (TP FP : \Real) : \Real := TP / (TP + FP)

/-- Recall (true positive rate) -/
def TPR (TP FN : \Real) : \Real := TP / (TP + FN)

/-- Fowlkes-Mallows index -/
def FM (TP FP FN : \Real) : \Real := \root (PPV TP FP * TPR TP FN)

/-- Matthews correlation coefficient -/
def MCC (TP TN FP FN : \Real) : \Real :=
  (TP * TN - FP * FN) / \root ((TP + FP) * (TP + FN) * (TN + FP) * (TN + FN))
\end{lstlisting}

Next, we define several lemmas that prove generic facts that will be useful in
the main theorem.

\begin{lstlisting}[language=lean, firstnumber=last]
/-- The basic fact `c / x \to 0` as `x \to +\Infinity`.
Limits in Lean are expressed with filters.
`Tendsto f atTop (\nhdsSym L)` is the Lean form of `lim_{x \to +\Infinity} f x = L`. -/
lemma tendsto_const_div_atTop_nhds_0 (c : \Real) :
    Tendsto (fun x : \Real => c / x) atTop (\nhdsSym 0) :=
  tendsto_const_nhds.div_atTop Filter.tendsto_id

/-- If `c/x \to 0` then `1 + c/x \to 1` (limit rule for addition). -/
lemma tendsto_one_add_const_div_atTop (c : \Real) :
    Tendsto (fun x : \Real => (1 : \Real) + c / x) atTop (\nhdsSym 1) := by
  simpa using (tendsto_const_nhds.add (tendsto_const_div_atTop_nhds_0 c))

/-- If `c/x \to 0` then `a - c/x \to a` (limit rule for subtraction). -/
lemma tendsto_const_sub_const_div_atTop (a c : \Real) :
    Tendsto (fun x : \Real => a - c / x) atTop (\nhdsSym a) := by
  simpa using (tendsto_const_nhds.sub (tendsto_const_div_atTop_nhds_0 c))

/- A common pattern we will use: `A * (1 + c/x) * (1 + d/x) \to A`.
This is just the limit rule for multiplication, plus the fact that constants tend to
themselves, and `(1 + c/x) \to 1`.-/
lemma tendsto_const_mul_one_add_mul_one_add_div_atTop (A c d : \Real) :
    Tendsto (fun x : \Real => A * (1 + c / x) * (1 + d / x)) atTop (\nhdsSym A) := by
  have := (tendsto_one_add_const_div_atTop c).mul (tendsto_one_add_const_div_atTop d)
  simpa [mul_assoc] using (tendsto_const_nhds.mul this)

/-- If `a > 0` then `sqrt(a) \ne 0` (since `sqrt(a) > 0`).
In Lean, a theorem or lemma is stated in the context of named hypotheses
(assumptions). Read this as: given the condition `h_agt0` the following claim is true. -/
lemma sqrt_of_pos_ne_zero {a : \Real} (h_agt0 : 0 < a) : \root a \ne 0 :=
  ne_of_gt (Real.sqrt_pos.mpr h_agt0)

/- A generic algebraic step used in the MCC denominator manipulation.
`sqrt(x) / t = sqrt(x / t^2)`  (assuming `0 \le x` and `0 \le t`).
`aesop` (Automated Extensible Search for Obvious Proofs) is an automation tactic.
It performs a small proof search using simp rules and standard lemmas. -/
lemma sqrt_div_eq_sqrt_div_sq {x t : \Real} (h_xge0 : 0 \le x) (h_tge0 : 0 \le t) :
    \root x / t = \root (x / (t ^ 2)) := by aesop
\end{lstlisting}

\newpage

We are now prepared to state the main theorem.
Under conditions where confusion matrix entries are non-negative, and both $\TP{} + \FP{}$ and $\TP{} +
  \FN{}$ are positive, the Matthews Correlation Coefficient converges towards the Fowlkes-Mallows index as
  true negatives become arbitrarily large:

\begin{lstlisting}[language=lean, firstnumber=last]
theorem tendsto_MCC_atTop_eq_FM
    {TP FP FN : \Real}
    (hTP_geq0 : 0 \le TP)
    (hFP_geq0 : 0 \le FP)
    (hFN_geq0 : 0 \le FN) 
    (hTPFPpos : 0 < TP + FP)
    (hTPFNpos : 0 < TP + FN) :
    Tendsto (fun TN : \Real => MCC TP TN FP FN) atTop (\nhdsSym (FM TP FP FN)) := by
\end{lstlisting}

With the claim now stated, the proof of this theorem is:

\begin{lstlisting}[language=lean, firstnumber=last]
  -- `A` is the constant factor that does not depend on TN.
  let A : \Real := (TP + FP) * (TP + FN)
  -- This is the "post step 3" expression: the same one that appears in the algebraic limit.
  let post_step3 : \Real \to \Real := fun TN =>
    (TP - FP * FN / TN) / \root (A * (1 + FP / TN) * (1 + FN / TN))
  ----------------------------------------------------------------------
  -- Step 1/2/3 algebraic rewrite: for TN > 0, MCC(TN) = post_step3(TN).
  ----------------------------------------------------------------------
  -- `f \eqF[atTop] g` means: `f TN` eventually equals `g TN` for all sufficiently large `TN`.
  have h_steps_123 :
      (fun TN : \Real => MCC TP TN FP FN) \eqF[atTop] post_step3 := by
    -- `filter_upwards` is a convenient way to work with "eventually" statements.
    -- Here it gives us an arbitrary `TN` with the assumption `0 < TN`.
    filter_upwards [Filter.eventually_gt_atTop (0 : \Real)] with TN hTN_gt0

    -- After unfolding, the goal is a concrete identity between real expressions.
    simp [MCC, post_step3, A]
    -- Step 2: distribute the factor `1/TN` into the numerator.
    have h_num : (TP * TN - FP * FN) / TN = TP - FP * FN / TN := by
      field_simp [hTN_gt0.ne'] 
    -- Step 3: rewrite `(TN + FP)/TN` as `1 + FP/TN`, similarly for `FN`.
    have h_inside :
        ((TP + FP) * (TP + FN) * (TN + FP) * (TN + FN)) / (TN ^ 2) =
          (TP + FP) * (TP + FN) * (1 + FP / TN) * (1 + FN / TN) := by field_simp [hTN_gt0.ne']
    -- Name the large product under the MCC square root.
    let mcc_inside_denom : \Real := (TP + FP) * (TP + FN) * (TN + FP) * (TN + FN)
    -- Side condition needed to move division under `sqrt`.
    have hDenomGe0 : 0 \le mcc_inside_denom := by simp [mcc_inside_denom]; positivity
    -- Denominator rewrite: push `/TN` inside `sqrt`, then substitute the Step-3 identity.
    have h_sqrt :
        \root ((TP + FP) * (TP + FN) * (TN + FP) * (TN + FN)) / TN =
          \root (A * (1 + FP / TN) * (1 + FN / TN)) := by
      simpa [mcc_inside_denom, A] using
        (sqrt_div_eq_sqrt_div_sq (x := mcc_inside_denom) (t := TN) hDenomGe0 hTN_gt0.le).trans
          (by aesop)
    -- Final algebraic combination: divide numerator and denominator by TN.
    calc
      (TP * TN - FP * FN) / \root ((TP + FP) * (TP + FN) * (TN + FP) * (TN + FN)) =
        ((TP * TN - FP * FN) / TN) /
          (\root ((TP + FP) * (TP + FN) * (TN + FP) * (TN + FN)) / TN) := by field_simp [hTN_gt0.ne']
      _ = (TP - FP * FN / TN) / \root (A * (1 + FP / TN) * (1 + FN / TN)) := by aesop
  ----------------------------------------------------------------------
  -- Limit of post_step3: as TN \to +\Infinity, the small fractions FP/TN and FN/TN go to 0.
  ----------------------------------------------------------------------
  -- Numerator limit: `TP - (FP*FN)/TN \to TP`.
  have h_num_lim :
      Tendsto (fun TN : \Real => TP - FP * FN / TN) atTop (\nhdsSym TP) :=
    tendsto_const_sub_const_div_atTop TP (FP * FN)
  -- Denominator limit: `sqrt(A * (1 + FP/TN) * (1 + FN/TN)) \to sqrt(A)`.
  have h_den_lim :
      Tendsto (fun TN : \Real => \root (A * (1 + FP / TN) * (1 + FN / TN))) atTop (\nhdsSym (\root A)) :=
      by simpa using
       (Filter.Tendsto.sqrt (tendsto_const_mul_one_add_mul_one_add_div_atTop A FP FN))
  -- The quotient limit rule needs the limit denominator to be nonzero.
  have h_den_ne : \root A \ne 0 := by
     have hApos : 0 < A := by simpa [A] using (mul_pos hTPFPpos hTPFNpos)
     exact sqrt_of_pos_ne_zero hApos
  -- Quotient limit rule: if num \to Num and den \to Den with Den \ne 0, then num/den \to Num/Den.
  have h_post_lim : Tendsto post_step3 atTop (\nhdsSym (TP / \root A)) := by
    exact Filter.Tendsto.div h_num_lim h_den_lim h_den_ne
  -- Transfer the limit from `post_step3` back to `MCC` using eventual equality.
  have h_mcc_lim :
      Tendsto (fun TN : \Real => MCC TP TN FP FN) atTop (\nhdsSym (TP / \root A)) := by
    exact h_post_lim.congr' (Filter.EventuallyEq.symm h_steps_123)
  ----------------------------------------------------------------------
  -- Rewrite FM into the same closed form `TP / sqrt(A)`.
  ----------------------------------------------------------------------
  have h_FM : FM TP FP FN = TP / \root A := by
    have: TP / (TP + FP) * (TP / (TP + FN)) = (TP ^ 2) / ((TP + FP) * (TP + FN)) := by
      field_simp [hTPFPpos.ne', hTPFNpos.ne']
    simp_all [FM, PPV, TPR, A]
  /- `h_mcc_lim` shows the limit is `TP / sqrt A`. `h_FM` shows `FM` can be rewritten as the same
  value. Substituting this rewrite into the target of `h_mcc_lim` completes the proof. -/
  simpa [h_FM] using h_mcc_lim
\end{lstlisting}

The above Lean code compiles with version
\texttt{leanprover/lean4:v4.28.0-rc1}, thus --- up to soundness holes in the
Lean kernel --- we have proven: $\lim_{\TN{} \to \infty} \MCC{} = \Fowlkes{}, \square{}$.

\section{LLM Usage}\label{sec:ai_usage}

The role of LLMs in scientific and mathematics research is growing~\cite{luo_llm4sr_survey_2025}.
In this work they played two important roles.
First, they helped the author, who was previously unfamiliar with Lean, to formalize this simple but
  non-trivial proof.
Second, they helped identify related work published in a different field with different terminology.
This section is anecdotal and is included as a record of how LLMs were used in this work.

\paragraph{Formalization with LLMs}

Lean is not a simple language.
Using it effectively requires a significant amount of background knowledge.
The author started making manual efforts to formalize this proof in 2023 without much progress.
Around this time LLMs such as GPT-4o~\cite{openai_gpt4o_system_card} and
  GPT-5~\cite{openai_gpt5_system_card} started demonstrating impressive capabilities.
However, the author's efforts to use them to help formalize the proof were unsuccessful.
The first successful version of the proof was written using
  GPT-5.1~\cite{openai_gpt51_system_card_addendum}.
It required significant manual feedback, was prone to syntax errors, and the result was 654 lines long, but
  it worked.
Further refinement from Aristotle~\cite{achim2025aristotleimolevelautomatedtheorem} and users on
  leanprover.zulipchat~\cite{crall2025zulip561031593} refined this to 66 lines (including comments).
Building on that, GPT-5.2~\cite{openai_gpt52_system_card} with extended thinking and manual feedback was
  able to produce the proof that better matches the structure of the original natural language argument, and
  is the version presented here.

\paragraph{Literature Search with LLMs}

When the original version of this paper (\url{https://arxiv.org/abs/2305.00594v1}) was posted to arXiv in
  2023, we were unaware of this limit relationship being noted elsewhere in the literature.
While revising the paper, we revisited the literature review using LLM-assisted search and identified that
  this result was known, but in a different context and with different names.
More details are given in \Cref{sec:related_work}.
We note this as an example of LLMs bridging a terminology gap in the search for related work.

\section{Conclusion}\label{sec:conclusion}

This paper proves that the limit of the MCC as the number of true negatives goes to infinity is the
  Fowlkes--Mallows index (\ie{} the geometric mean of precision and recall) and provides a Lean 4
  formalization.
We interpret this result in the context of object detection where the number of true negatives is
  effectively unbounded.
We connect this to a previous observation in the ecology literature and briefly discuss the role of LLMs in
  finding that connection.

\section{Acknowledgements}\label{sec:ack}

Thanks to Lee Newberg for the cleaner formulation of \cref{eq:lee_step1,eq:lee_step2,eq:lee_step3}.

\bibliographystyle{unsrtnat}
\bibliography{references}  

\end{document}